\documentclass[preprint,12pt,authoryear]{elsarticle}
\usepackage{natbib}  
\usepackage{amssymb}
\usepackage{amsmath,amsfonts}
\usepackage{algorithmic}
\usepackage{graphicx}
\usepackage{textcomp}
\usepackage{xcolor}
\usepackage{ulem}
\usepackage{subfigure}
\usepackage{diagbox}
\usepackage{pifont}
\newcommand{\cxmark}{\ding{55}}
\usepackage{threeparttable} 
\usepackage{multirow}
\def\BibTeX{{\rm B\kern-.05em{\sc i\kern-.025em b}\kern-.08em
    T\kern-.1667em\lower.7ex\hbox{E}\kern-.125emX}}
\usepackage{pdfpages}

\journal{ }

\begin{document}

\includepdf[pages=1]{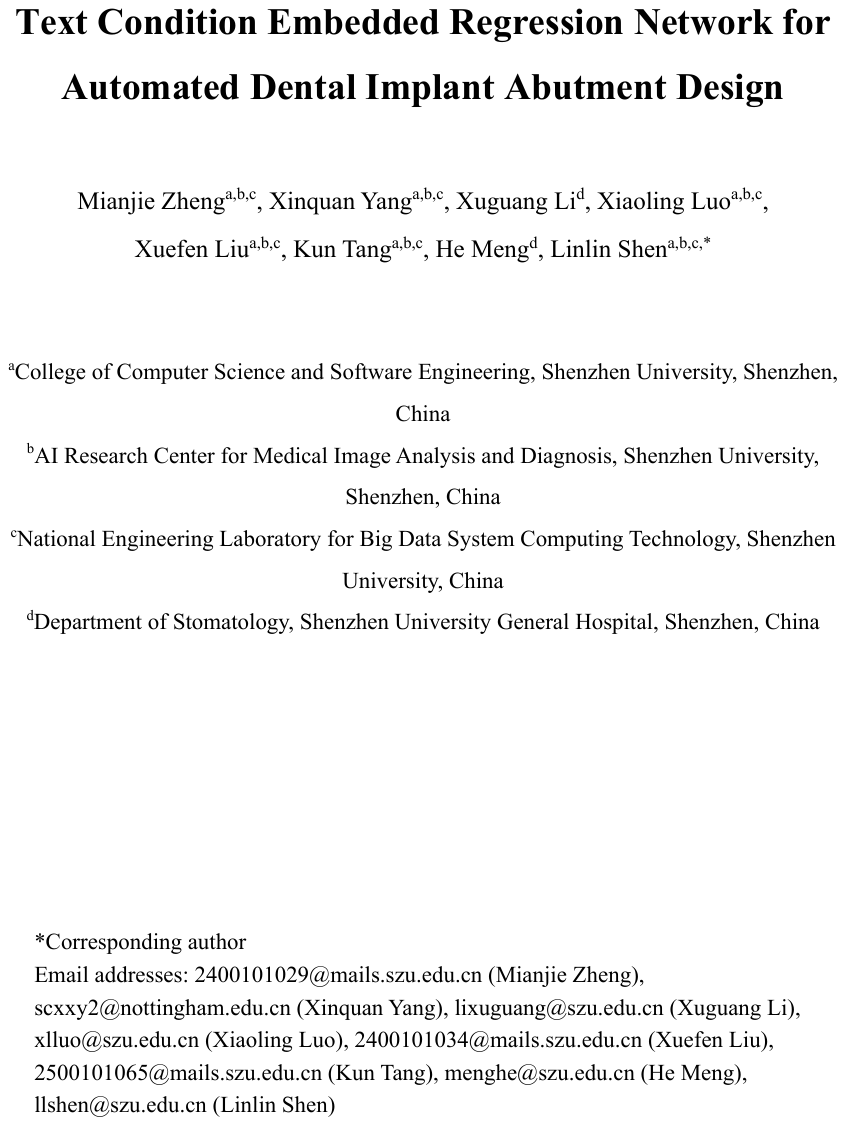}

\begin{frontmatter}
\title{Text Condition Embedded Regression Network for Automated Dental Implant Abutment Design}

\author[label1,label2,label3]{Mianjie Zheng}
\ead{2400101029@mails.szu.edu.cn}
\author[label1,label2,label3]{Xinquan Yang}
\ead{scxxy2@nottingham.edu.cn}
\author[label4]{Xuguang Li}
\ead{lixuguang@szu.edu.cn}
\author[label1,label2,label3]{Xiaoling Luo}
\ead{xlluo@szu.edu.cn}
\author[label1,label2,label3]{Xuefen Liu}
\ead{2400101034@mails.szu.edu.cn}
\author[label1,label2,label3]{Kun Tang}
\ead{2500101065@mails.szu.edu.cn}
\author[label4]{He Meng}
\ead{menghe@szu.edu.cn}
\author[label1,label2,label3]{Linlin~Shen\corref{mycorrespondingauthor}}
\ead{llshen@szu.edu.cn}
\cortext[mycorrespondingauthor]{Corresponding author}

\address[label1]{College of Computer Science and Software Engineering, Shenzhen University, Shenzhen, China}
\address[label2]{AI Research Center for Medical Image Analysis and Diagnosis, Shenzhen University, Shenzhen, China}
\address[label3]{National Engineering Laboratory for Big Data System Computing Technology, Shenzhen University, China}
\address[label4]{Department of Stomatology, Shenzhen University General Hospital, Shenzhen, China}

\begin{abstract}
The abutment is an important part of artificial dental implants, whose design process is time-consuming and labor-intensive. Long-term use of inappropriate dental implant abutments may result in implant complications, including peri-implantitis.
Using artificial intelligence to assist dental implant abutment design can quickly improve the efficiency of abutment design and enhance abutment adaptability. 
In this paper, we propose a text condition embedded abutment design
framework (TCEAD), the novel automated abutment design solution available in literature. 
The proposed study extends the self-supervised learning framework of the mesh mask autoencoder (MeshMAE) by introducing a text-guided localization (TGL) module to facilitate abutment area localization.
As the parameter determination of the abutment is heavily dependent on local fine-grained features (the width and height of the implant and the distance to the opposing tooth), we pre-train the encoder using oral scan data to improve the model's feature extraction ability. Moreover, considering that the abutment area is only a small part of the oral scan data, we designed a TGL module, which introduces the description of the abutment area through the text encoder of Contrastive Language-Image Pre-training (CLIP), enabling the network to quickly locate the abutment area. 
We validated the performance of TCEAD on a large abutment design dataset. 
Extensive experiments demonstrate that TCEAD achieves an Intersection over Union (IoU) improvement of 0.8\%–12.85\% over other mainstream methods, underscoring its potential in automated dental abutment design.
\end{abstract}
\end{frontmatter}
\section{Introduction}\label{introduction}
Dental implant restoration is the mainstream method for treating dental arch defects and tooth loss \citep{benakatti2021dental, yang2024two}, in which dental implant restoration treatment involves two processes: implant-supported restoration placement and fabrication of the implant-supported restoration. 
The first step in fabricating the implant-supported restoration is to select the appropriate dental implant abutment. 
The dental implant abutment configuration must consider multiple clinical parameters, including the dental implant model of the patient, the gingival contour, and the occlusal space, which directly influence both the functional and biological outcomes of the restoration. Improper abutment design may lead to unfavorable stress distribution or biological complications such as soft tissue inflammation and peri-implantitis, ultimately compromising long-term implant stability \citep{albakri2024mechanical}.
Traditionally, the abutment was designed by measuring the restorative space for the implant abutment manually or using CAD-CAM methods \citep{kim2012fabrication}.
Since the internal space of each person's mouth is different, the restoration space for dental implants will also vary, which will affect the design of the customised abutment. 
Specifically, the measurement of the restoration space mainly involves measuring the thickness of the oral gingiva (transgingival), the diameter of the implant position (diameter), and the gingival-mandibular distance (height). 
In addition, traditional methods require scanning and printing the 3D structure of the oral cavity, followed by manual measurement of the restoration space.
In contrast, the CAD-CAM method involves only scanning the 3D structure of the oral cavity and then using CAD software to measure it on a computer \citep{tartea2023comparative, benakatti2021dental}. 
In general, these manual methods are time-consuming and labor-intensive, requiring extensive clinical experience, and inappropriate abutment selection may lead to biological complications.
In recent years, artificial intelligence (AI) has achieved great success in the dental field (tooth segmentation, crown restoration). 
Therefore, there is great potential to use artificial intelligence to automate the design of abutments.

\begin{figure*}[htbp]
\centering
\includegraphics[scale=0.24]{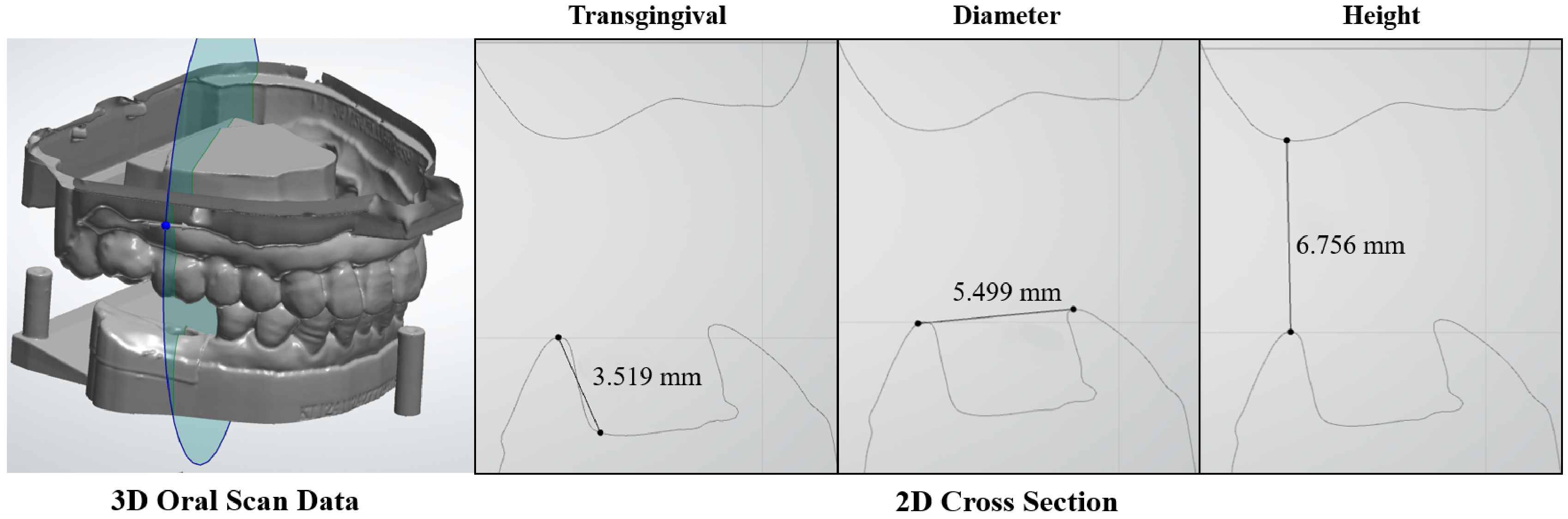}
\caption{The process of manually determining abutment parameters. The blue part in 3D oral scan data represents the position of the cross-section, and the three 2D cross-sections represent the measurements of the transgingival, diameter and height, respectively.}
\label{fig1}
\end{figure*}

\par
The oral scan data is the basis for abutment design, which mainly consists of 3D point clouds and mesh data. Recently, many neural networks have been proposed and applied to such data. 
Qi et al. proposed PointNet~\citep{qi2017pointnet} and PointNet++ \citep{qi2017pointnet++} to process unordered point sets, which learns multiple-scale features to obtain local structural information. 
Masci et al. designed a MeshCNN to process 3D mesh data~\citep{hanocka2019meshcnn}, defining sorting-invariant convolution operations to overcome the variable number of vertices in the mesh.  
In the field of dentistry, existing works mainly focus on the tooth segmentation tasks~\citep{cui2021tsegnet,jang2021fully,qiu2022darch,xu20183d} and crown restoration~\citep {tian2021efficient,shen2023transdfnet,yuan2020personalized}.
However, these segmentation and generation networks are not suitable for automated abutment design. They focus more on the surface of the teeth (e.g., tooth size and curvature). 
Different from these tasks, the abutment design needs the spatial information of the implant, including its depth, diameter, and distance from the opposing teeth. 
As shown in Fig.~\ref{fig1}, the abutment restoration space is determined by measuring the cross-section of the implanted tooth, i.e., the values of transgingival, diameter, and height. 
Therefore, designing a regression network specific to the prediction of abutment parameters is necessary.

\begin{figure}[htbp]
\centering
\includegraphics[width=0.8\linewidth]{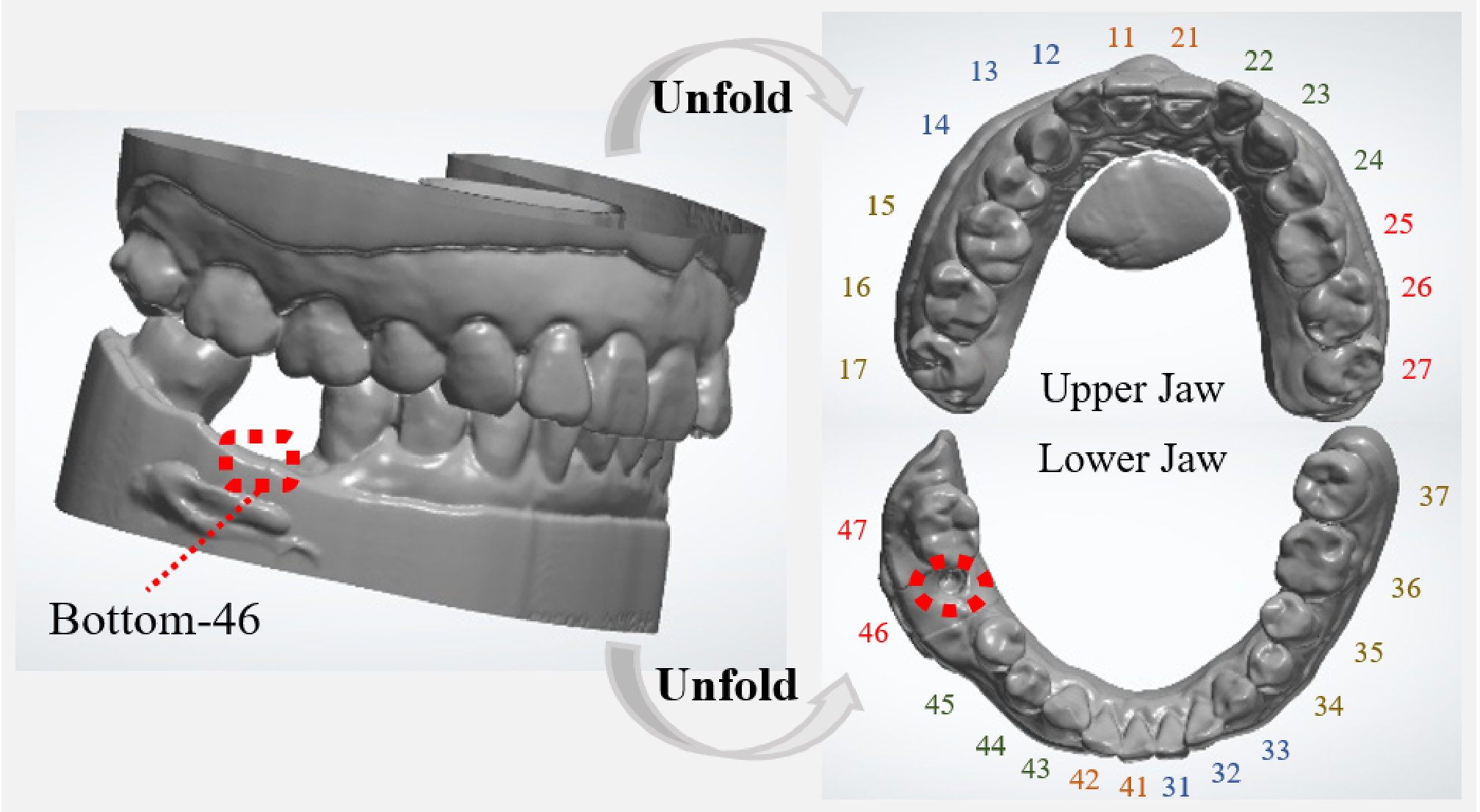}
\caption{The tooth index according to the FDI World Dental Federation numbering system. 
}
\label{TeethDetail}
\end{figure}

In this paper, we designed the novel automated abutment design framework (TCEAD), which takes the oral scan data as input and directly outputs the parameters of the abutment, greatly simplifying the abutment design process. 
The architecture of our framework is given in Fig.~\ref{overall_framework}. It consists of a pre-training and fine-tuning stage. 
In the pre-training stage, we introduced a masked autoencoder architecture and used a large amount of oral scan data for pre-training, aiming to enhance the network's ability to capture the fine-grained oral features, thereby facilitate the regression task of the abutment parameter.
During the fine-tuning stage, the pre-trained encoder is employed to extract features from the input oral data. 
Fig.~\ref{TeethDetail} shows the tooth index according to the FDI World Dental Federation numbering system.
Clinically, dentists will mark the tooth index of the implant area and input such information to the implant system before designing the abutment. 
This information can help the network quickly locate the implant area. 
Considering that segmenting this area requires additional manual annotation, we propose to use a simpler text-guided method. Specifically, we designed a text-guided localization (TGL) module, which takes the implant area location information as prior to guide the prediction network. TGL first extracts text embedding from the Contrastive Language-Image Pre-training (CLIP)~\citep{radford2021learning} according to the description of the teeth position. 
Then, we combine text embedding and the network features through cross-modal interaction, promoting the model to focus on the area of abutment. 
We validated the performance of our framework on a large abutment design dataset. The experimental results show that the abutments designed by our framework meet clinical needs and greatly improve the design efficiency.
In summary, the contributions of this paper are as follows:
\begin{itemize}
    \item We propose a text condition embedded abutment design framework (TCEAD), the first automated implant abutment design solution, to automate greatly simplifying the abutment design process.
    \item A text-guided localization (TGL) module was designed to introduce the description from the dentist to guide the model to focus on the area of abutment.
    \item Extensive experiments were conducted on a large abutment design dataset, the results show that the abutments designed by TCEAD meet clinical needs and greatly improve the design efficiency.
\end{itemize}

\section{Related Work}
\subsection{Traditional Abutment Design}
Measuring the restoration space of the abutment is a critical step in pre-selecting abutments \citep{abichandani2013abutment}. Previously, manual measurements were required to determine the restoration space. 
The restoration space consists of the thickness of the oral gingiva (transgingival), the diameter of the implant position (diameter), and the gingival-mandibular distance (height). 
On the other hand, the process of manually preselecting abutments is extremely cumbersome \citep{shah2023literature}. 
It requires scanning the internal structure of the mouth, printing the gums, measuring the restoration space, preselecting abutments, installing the abutments to determine whether they are suitable, and finally preselecting the abutments.
Obviously, this method is not efficient for the mass production of customized abutments, and the probability of complications due to mismatch will also be greatly increased.
To solve this problem, more and more abutments are being manufactured using CAD measurements on computers \citep{kim2012fabrication}. CAD-CAM technology customised abutments offer better fit and higher personalisation, avoiding the dimensional inaccuracies caused by traditional waxing, investing and casting steps \citep{zhang2017method}. 
This greatly reduces material waste caused by manual selection of abutments and shortens the abutment selection process \citep{torres2009cad}. 
However, these methods still have limitations. Manual or computer measurements require locating and measuring the restoration space, and some data deviations may occur during measurement, which need to be handled by experts. 
At the same time, design errors introduced during manual abutment fabrication may compromise long-term implant stability and lead to biological complications.

\subsection{Deep Learning Method Based on Mesh Data}
Point cloud and mesh are common three-dimensional data. Generally, the point clouds only consider the connections between points, making it difficult to represent the structural information.
On the contrary, mesh data includes the intrinsic geometric structure, e.g., vertices, areas, and face normals. 
Therefore, researchers prefer to use mesh data to handle tasks that require fine-grained features.
However, the irregular mesh structure and disordered faces in mesh data make great challenges. 
Boscaini et al. attempted to use convolutional networks to process complex mesh data, and their experiments proved the effectiveness of the method~\citep{boscaini2015learning}. 
Feng et al. proposed MeshNet to address the complexity and irregularity of meshes by integrating surface elements and triangular network features~\citep{feng2019meshnet}. 
To better represent the surface and topological structure of polygonal meshes, Hanocka et al. proposed MeshCNN~\citep{hanocka2019meshcnn}, which utilises edge folding technology to perform convolution and pooling, thereby learning important features of the mesh.
Hu et al. designed a subdivnet\citep{hu2022subdivision} that hierarchically processes the mesh from fine to coarse, re-regularising the mesh to facilitate network learning. 
Singh et al. developed a MeshNet++~\citep{singh2021meshnet++} that can learn local features of mesh at multiple scales to improve recognition accuracy. 
With the development of the transformer in the field of computer vision, researchers began to develop transformer-based mesh networks.
Liang et al. proposed MeshMAE~\citep{liang2022meshmae} as a powerful mesh pre-training framework, which enhances the network's understanding of 3D mesh data through a large amount of 3D mesh pre-training. 
Li et al.~\citep{li2022laplacian} proposed a Laplacian mesh transformer that can effectively extract key structure and collective features from mesh data, and capture local information of the mesh through a dual attention mechanism. 

\begin{figure*}\centering
\includegraphics[scale=0.27]{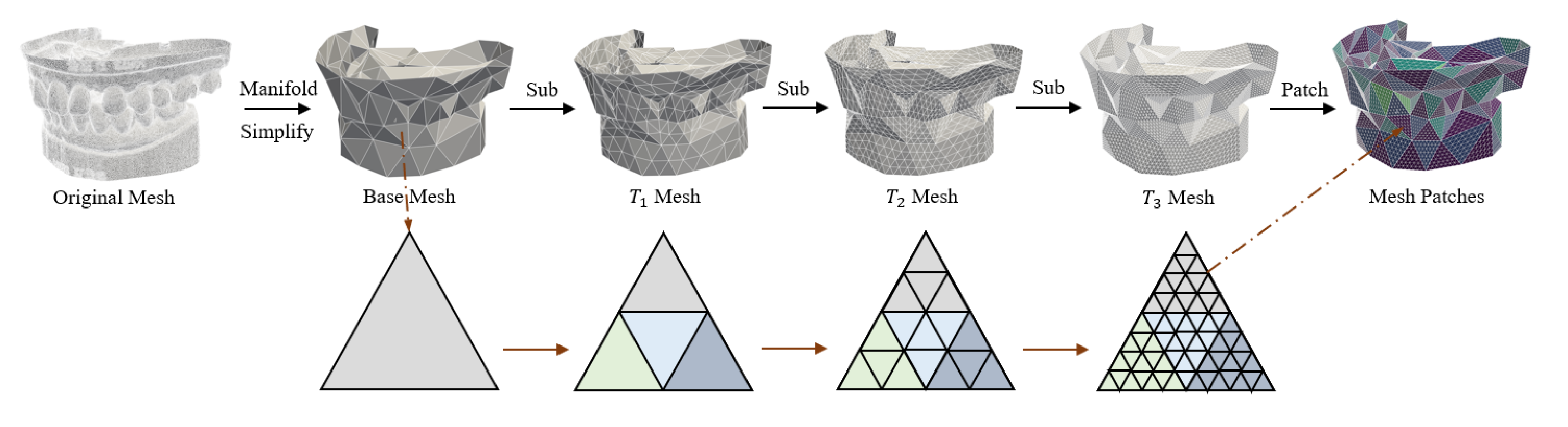}
\caption{Visualization of the remesh processing. First, perform manifold and simplification operations on the original mesh to process the oral scan data into a base mesh with 500 standard faces. Furthermore, refine the base mesh three times, each time refining one face into four faces. Finally, the refined mesh is projected as input. The refined mesh is divided into 500 patches, each of which contains 64 faces arranged in a fixed order after remesh processing.}
\label{Remeshpro}
\end{figure*}

\subsection{Mesh Data in Dentistry}
Mesh data also plays a crucial role in the field of dentistry. 
Lian et al. proposed a MeshSegNet~\citep{lian2020deep} for automatic labeling of 3D tooth surfaces. By hierarchically learning multiscale features and integrating local and global information to achieve automatic labeling. 
Xu et al.~\citep{xu20183d} extracted a set of geometry features as face feature representations and produced a probability vector to indicate the probability that a face belongs to the corresponding model part, greatly improving the efficiency of the feature extraction stage. 
He et al.~\citep{he2022unsupervised} proposed a method that combined unsupervised pretraining with supervised fine-tuning, which achieved higher segmentation accuracy on a small amount of labeled data. 
Wang et al. designed a weakly supervised tooth instance segmentation network (WS-TIS) that can achieve accurate tooth instance segmentation with only subject-level class labels and approximately 50\% point-by-point tooth annotations~\citep{wang2024weakly}. 
Liu et al. utilised the prompt segmentation capability of the Segment Anything Model (SAM) to supplement sparse supervisory information~\citep{liu20243d}. 
Huang et al. used multi-view 2D border annotations on 3D mesh data and designed a multi-view prompt-driven tooth segmentation method based on SAM~\citep{huang2024iossam}. 
Almalki et al. designed a self-supervised learning DentalMAE~\citep{almalki2024self}, which uses the predicted embeddings of masked mesh triangles as the basis for loss evaluation.
In the task of dental crown reconstruction, Hosseinimanesh et al. proposed dental mesh completion (DMC)~\citep{hosseinimanesh2023mesh}, which uses an encoder-decoder architecture and a differentiable point-to-mesh layer to directly generate dental crown meshes based on a point cloud context. 
Yang et al. designed a point-to-mesh generation network with a morphology-aware cross-attention module, which generates crown meshes directly from 3D scan points~\citep{yang2024dcrownformer}. 
Hosseinimanesh et al.~\citep{hosseinimanesh2025personalized} designed an end-to-end deep learning model that automatically generates personalised crown meshes using point cloud information from the prepared tooth, adjacent teeth, and the two closest teeth on the opposite jaw. 

\begin{figure*}[htbp]
\centering
\includegraphics[scale=0.31]{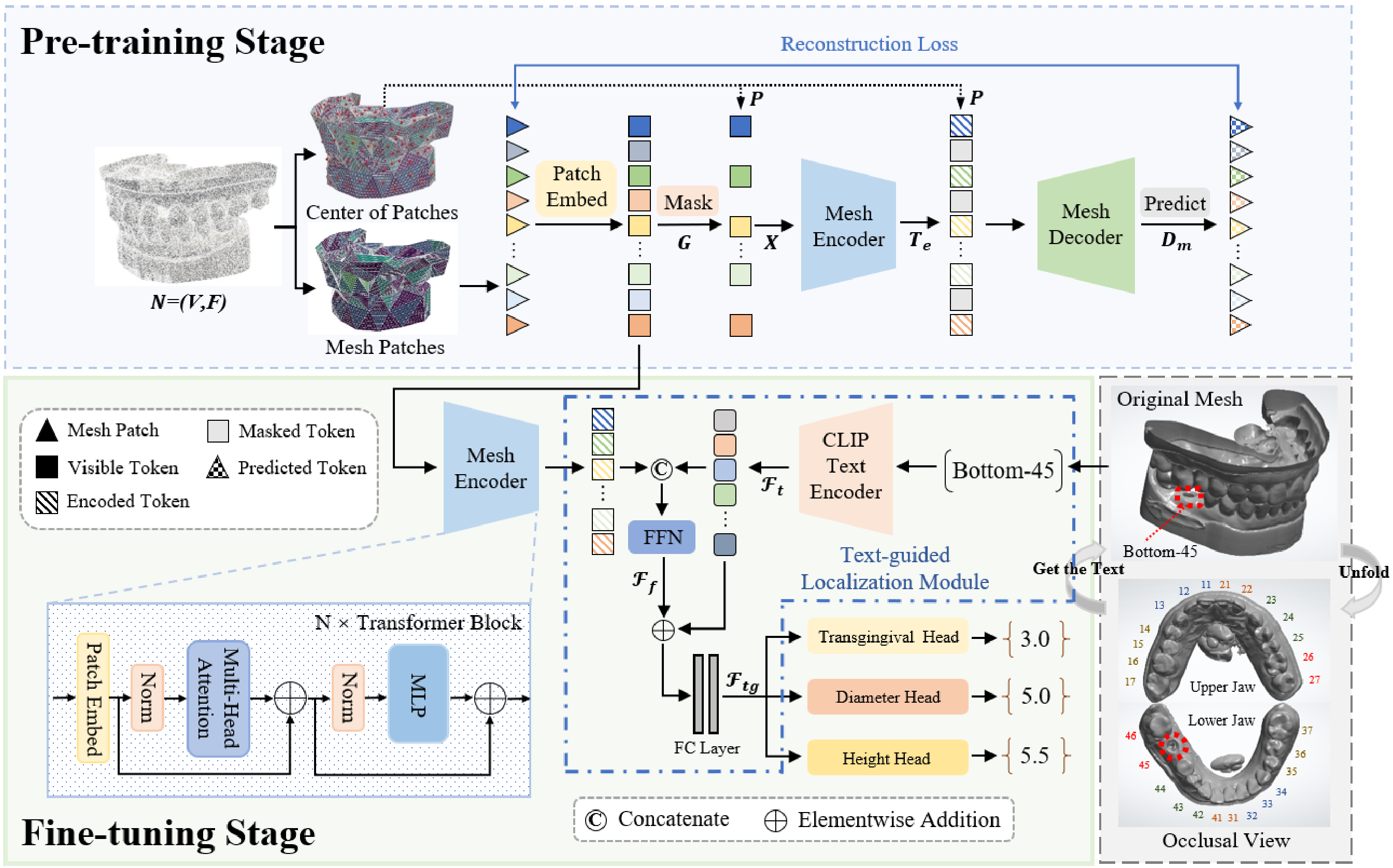}
\caption{The overall framework of the proposed TCEAD method. The input mesh is first divided into multiple non-overlapping patches through the remeshing and then embedded through an MLP. In the pre-training stage, a random subset of patches is masked, and only the visible patches are processed by the mesh encoder, while the decoder reconstructs the original input using both the encoded visible embeddings and the masked embeddings. During the fine-tuning stage, the pre-trained weights are transferred to the network to support downstream tasks. Subsequently, the text-guided localization (TGL) module fuses the abutment position description with the mesh features and feeds them to the regression head. Finally, the regression head predicts the abutment parameters.
}
\label{overall_framework}
\end{figure*}

\section{Method}

\subsection{Oral Scan Data Pre-training}
Fig.~\ref{fig1} shows the process of determining the abutment parameter, which heavily relies on the cross-section of the implant area, i.e., the depth and diameter of the implant area, and distance from the opposing teeth. 
Therefore, the network is required to capture fine-grained features from oral scan data. 
To this end, we choose to pre-train the encoder using a large amount of oral scan data, which will enhance the model's ability to extract features. 
However, unlike the division of conventional patches of 2D images, a mesh consists of unordered vertices and triangular mesh faces in 3D space, defined as $N = (V, F)$. 
To be effectively applied to the masked autoencoder, based on the properties of the triangular face, the face features are divided into 13-dim vectors involving area, three face normals, three interior angles of the triangular face, three coordinates of the center point of the face, and the interior product of the multiplication of the face normals and vertex normals. 
Based on the above operations, we process patches in two stages: mesh patch generation and patch embedding.

\textbf{Mesh Patch Generation.}
Unlike traditional image inputs, mesh has irregular and disordered properties, and it is not reasonable to directly classify multiple disordered faces into patches. 
Following subdivnet~\citep{hu2022subdivision}, the MAPS algorithm is applied to remesh the original mesh to divide the mesh data into structural and hierarchical rules. 
The remesh process involves simplification and multiple subdivisions, each of which subdivides the mesh of $M$ faces into a mesh of $4M$ faces ($496 \leq M \leq 500$ in our experiments). 
Specifically, each original mesh is first reduced to a base mesh of $M$ faces. The base mesh is then subdivided $T$ times to finally obtain a mesh of $4^TM$ faces. 
Thus, the final mesh is divided into $M$ patches, where faces are arranged in a consistent order in each patch and 10-dimensional feature vectors of all faces are concatenated to form a uniform feature representation of the patch.

\textbf{Patch Embedding and Masking.}
After the remesh process, the 13-dimensional feature vectors of all faces belonging to the same patch are connected in an ordered manner and have predictable properties. 
The representation of each patch can be mapped to $\{g_i\}^m_{i=1}$ by an MLP, where $m$ denotes the number of patches. 
The positional information of each patch is represented by the center coordinates of the constituent faces. 
Specifically, we use an MLP layer to map the center coordinates of the faces in each patch to obtain positional embeddings of $\{p_i\}^m_{i=1}$. 
The final input is expressed as an embedding sequence $X=\{x_i\}^m_{i=1}$ consisting of patch embeddings $G=\{g_i\}^m_{i=1}$ and positional embeddings $P=\{p_i\}^m_{i=1}$.

Furthermore, patch embeddings $G$ are randomly masked using a masking rate $r$. 
The masked embeddings $G_m$ are replaced with a shared learnable masked embedding $G_l$ whose positional embedding remains unchanged and is finally fed to the encoders.

\subsubsection{\textbf{Encoder and Decoder}}
Our pre-training network is an encoder-decoder architecture. The architecture is given in Fig. \ref{overall_framework}. 
The encoder consists of 12 standard transformer blocks, including the visible input tokens $T_v$ and the positional embeddings added to each block. 
The encoded token is denoted as $T_e$, and the formula is expressed as follows:
\begin{equation}
	T_e = Encoder(T_v), \ \ \ T_e\in\mathbb{R}^{(1-r)m\times d},
\end{equation}
where $m$ denotes the number of patches, $r$ is the masking ratio, and $d$ is the embedding dimension. 
The decoder consists of 6 standard transformer blocks. Different from the encoder, the decoder takes the shared mask embedding to predict the missing patches. 
The visible embeddings and mask embeddings are jointly fed into the decoder, and all positional embeddings are added to all embeddings to provide position information for all the patches. 
Finally, the decoder outputs the decoded mask tokon $D_m$. The decoder structure can be expressed by the following equation:
\begin{equation}
	D_m=Decoder(concat(T_e, T_m)), \ \ \ D_m\in\mathbb{R}^{rm\times d},
\end{equation}
where $T_m$ denotes masked tokens.

\subsubsection{\textbf{Reconstruction Targets}}
The reconstruction target is defined as reconstructing the shape of the masked patches, including the vertices of the patches and the face features of the patches. 
For the 45 vertices of each masked patch, the coordinates relative to the center point of the patch are expressed as $c=(x, y, z)$. 
The output coordinates of the decoder are defined as $C_p=\{c_{pi}\}_{i=1}^{45}$, and the coordinates of the ground truth vertices in the patch are denoted as $C_g=\{c_{gi}\}_{i=1}^{45}$. 
We calculate $C_p$ and $C_g$ using the $L^{2}$-form chamfer distance:
\begin{equation}
\begin{split}
L_{CD}(C_{p}, C_{g}) = &\frac{1}{|C_p|}\sum_{p \in C_p}\min_{g \in C_g} ||p - g|| \\
&+\frac{1}{|C_g|}\sum_{g \in G_r} \min_{p \in C_p} ||g - p||,
\end{split}
\end{equation}

Moreover, due to the characteristics of mesh data, it is difficult to recover the geometric architecture of the entire mesh data by relying only on point features. Therefore, all face features for each patch need to be reconstructed and evaluated for loss.
To this end, we add a linear layer after the decoder to predict the face features of each patch, and the $L_{MSE}$ loss is used to evaluate the effect of the reconstructed face features. Therefore, the reconstruction loss consists of $L_{CD}$ and $L_{MSE}$, and the formula is expressed as follows:
\begin{equation}
L=L_{CD}+\varphi L_{MSE},
\end{equation}
where $\varphi$ denotes the loss weight of the face features.
	\par

\subsection{Text Condition Embedded Abutment Design Framework}
\subsubsection{\textbf{Text-guided localization Module}}
In clinical practice, dentists will mark the location of the implant area and specify the implant system before designing the abutment. This information can be used as a prior to guide the network to focus on the implant area.
Researchers usually segment planting areas based on location information, but this requires a lot of manual labeling.
Unlike segmentation labels, it is simple to generate corresponding text using existing location information, and existing works~\citep{yang2023tceip,yang2023tcslot} demonstrate that the effectiveness of introducing the text to guide the network to focus on the implant region.
Therefore, we designed a text-guided localization (TGL) module.
The architecture of TGL is given in Fig.~\ref{overall_framework}.
Specifically, we use the encoder of CLIP to generate text embedding $\mathcal{F}_t$ according to the description of the teeth position $I_{text}$:
\begin{equation}
	\mathcal{F}_t = CLIP(I_{text}),  \ \ \ \mathcal{F}_t\in\mathbb{R}^{1\times W},
\end{equation}
where $I_{text}$ is a fixed sentence, such as ‘This is a medical image of the missing top/bottom $j$-th tooth’. 

Since the text embedding and network features have a great difference. To strengthen the interaction between both modal features, we concatenate both features and then map them through the feedforward network (FFN) to induce the feature fusion:
\begin{equation}
	\mathcal{F}_{f} =\mathcal{Y} (concat(\mathcal{F}_t, \mathcal{F}_m)),  \ \ \ \mathcal{F}_{f}\in\mathbb{R}^{1\times W},
\end{equation}
where $\mathcal{F}_m \in \mathbb{R}^{1\times D}$ is the input encoder embeddings obtained by the max pooling operation and $\mathcal{Y}$ denotes FFN. In the end, we add text features to the fused features to strengthen the guidance of text embedding:
\begin{equation}
	\mathcal{F}_{tg} =\mathcal{Y} (\mathcal{F}_{fusion} + \mathcal{F}_t),  \ \ \ \mathcal{F}_{tg}\in\mathbb{R}^{1\times H}.
\end{equation}

\subsubsection{\textbf{Encoder and Decoder}}
In the fine-tuning stage, we use the pre-trained encoder as a feature extractor. 
The decoder consists of two fully connected layers and three independent regression heads. The fully connected layer first reduces the dimensionality of the fused features and then inputs them into the regression head. We combine mean squared error (MSE) loss and smooth L1 loss to supervise the regression heads. The total loss is expressed as:
\begin{equation}
	L_{1}(z_i)=
	\begin{cases} 
	\dfrac{1}{2} z_i^2, & \lVert z_i \rVert \leq \mathfrak{h}, \\
	\mathfrak{h} \cdot \lVert x \rVert - \dfrac{1}{2} z_i^2, & \lVert z_i \rVert > \mathfrak{h},
	\end{cases}
\end{equation}
\begin{equation}
	L_{total} = L_{MSE}^{\ddagger} + L_{1},
\end{equation}
where $z_i$ represents the difference between the i-th predicted value and the ground-truth, and $\mathfrak{h}$ is a smoothing function with a default value of $1$.

\section{Experiments}
\subsection{Datasets}
We collect a large abutment design dataset to evaluate the effectiveness of our method, the distribution of the dataset is given in Table~\ref{Data_num}. 
The total number of abutment design data is 6773, and we randomly divide it into a training set of 5494 and a test set of 1279. 
Moreover, there are 1371 oral scan data unlabeled data, i.e. without annotation of abutment design parameters.
For the pre-training stage, we use a publicly available Teeth3Ds dataset~\citep{ben2022teeth3ds+}, the training set and unlabeled data for training.

\begin{table}[htbp]
    \centering
        \caption{The dataset distribution in our experiment.}
        \begin{tabular}{c|c|c|c}
        \hline
        \multicolumn{2}{c|}{Pre-training} & \multicolumn{2}{c}{Fine-tuning} \\ \hline
        Dataset& Number& Split & Number  \\ \hline
        Teeth3DS  & 932 & Train & 5494 \\ 
        Collected Data  & 5494 + 1371 = 6865  & Test & 1279 \\ \hline
        \end{tabular}
\label{Data_num}
\end{table}

Since the original mesh data contained more than 200,000 faces, it required heavy computational resources.
Therefore, we use manifold simplification and remeshing to preprocess irregular oral mesh data.
Specifically, we simplified each mesh to 500 faces while repairing non-watertight areas. 
Then, we use the MAPS algorithm~\citep{liu2020neural} to re-regularize all meshes, subdivide each mesh three times, and finally obtain a mesh containing 32,000 faces. 
The refinement process is shown in Fig.~\ref{Remeshpro}. 
Each refinement divides 1 face into 4 faces, and three refinements divide it into 64 faces.

\subsection{Evaluation Criteria}
In clinical practice, dentists first measure the abutment implant area to obtain specific parameter values. They then select a finished abutment with the closest parameter values according to the implant system. Therefore, we used the parameter values of the finished abutment as the ground-truth design and compared them with those predicted by the proposed TCEAD method. To account for the deviation between the actual measured values and the ground-truth values, we adopted the Intersection over Union (IoU) metric to evaluate the accuracy of the predicted results.
We treat each abutment parameter as an area with a fixed size of one unit, and evaluate the results by calculating the IoU between the predicted value and the ground-truth value. Specifically, when the predicted value is denoted as $x$ and the corresponding ground-truth value is $y$, the intersection region is defined as $pv_i\cap gt_i$, where $pv_i$ represents the 1-unit range with the predicted value as the lower limit (x+1), and $gt_i$ represents the 1-unit range with the ground-truth value as the lower limit (y+1).
The IoU formula is expressed as follows:
\begin{equation}
	IoU(pv_i, gt_i) = \frac{|pv_i\cap gt_i|}{|pv_i\cup gt_i|}.
\end{equation}
When the IoU is 0.42, the difference between the predicted value and the ground-truth value is only 0.4 mm, which basically satisfies the manufacturing tolerance requirements.

\subsection{Implementation Details}
Pytorch is used for model pre-training and fine-tuning. 
For the pre-training of TCEAD, we use a batch size of 128, AdamW optimizer, and a learning rate of 1e-4 with a cosine learning schedule. 
Three data augmentation methods, i.e., random scale, random rotation, and shape deformation, are employed. 
The network is pre-trained for 300 epochs. 
For the fine-tuning of TCEAD, we use a batch size of 64, AdamW optimizer, and a learning rate of 1e-4 for network training. 
The data augmentation methods are same as the pre-training stage. 
The network is trained for 100 epochs, and the learning rate is decayed by a factor of 0.1 at 30 and 60 epochs, respectively. 
All the models are trained and tested on the platform of the NVIDIA A40 GPU.

\begin{figure*}[htbp]
    \centering
    \subfigure{}{\includegraphics[width=0.32\textwidth]{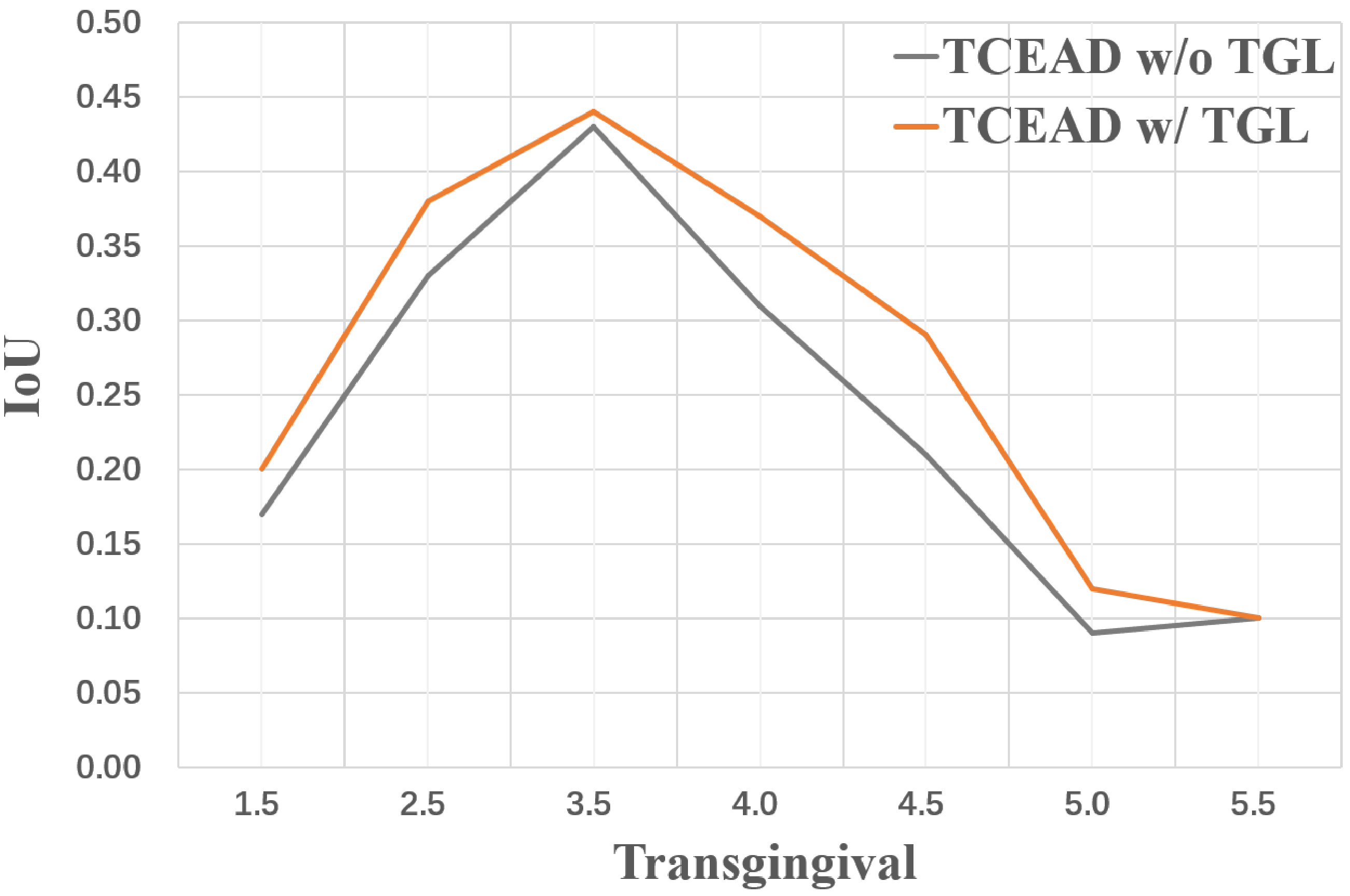}}
    \subfigure{}{\includegraphics[width=0.32\textwidth]{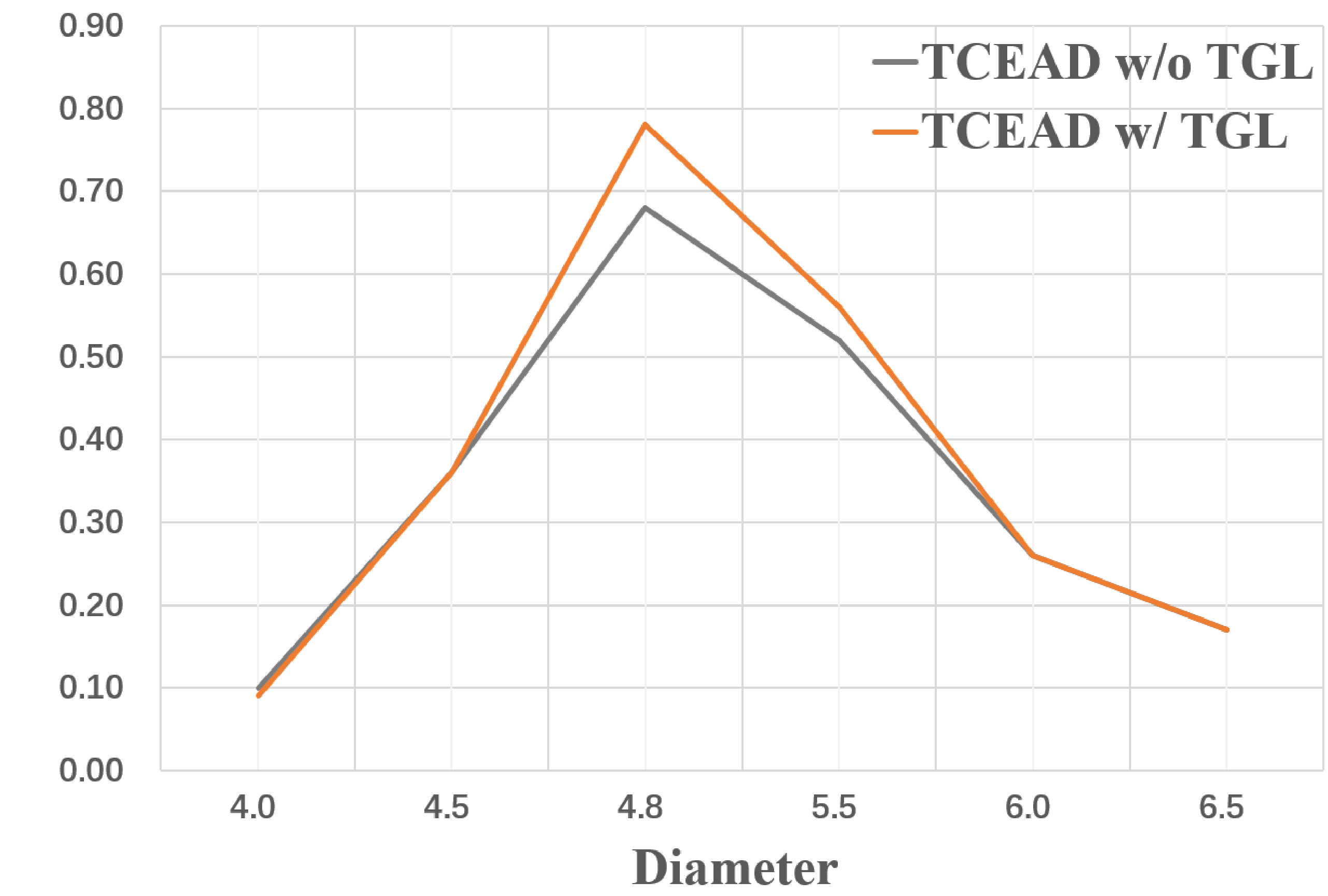}}
    \subfigure{}{\includegraphics[width=0.32\textwidth]{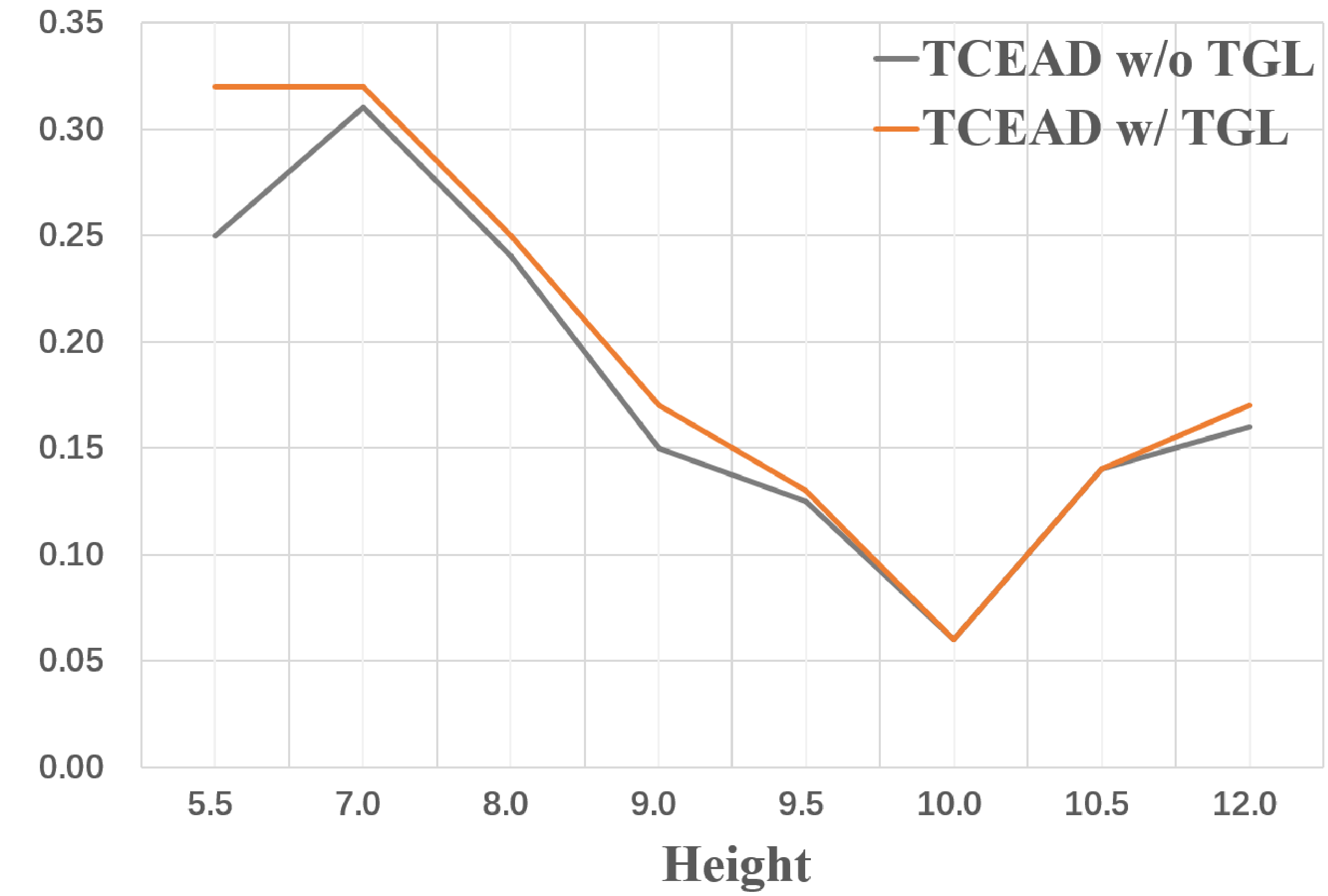}}
    \caption{Ablation experiment of the TGL component. The vertical axis represents the IoU performance, while the horizontal axis corresponds to value of each abutment parameter.}
    \label{Compare_TLM}
\end{figure*}

\subsection{Ablation Studies}
\subsubsection{Ablation Study of TGL}
To validate the effectiveness of the proposed text-guided localization (TGL) module, we conducted experiments on TCEAD with and without TGL. 
The experimental results are shown in Fig.~\ref{Compare_TLM}. 
As can be seen from Fig.~\ref{Compare_TLM}, TECAD with TGL shows better IoU for multiple parameters. 
Although the introduction of TGL does not improve performance in some numerical values, this is due to the small amount of data in this part. 
These experimental results prove the robust performance of TGL in different oral scan data.

\subsubsection{Data Type for Pre-training}
The collected oral scan data for network pre-training not only includes occlusal data that combines the upper and lower jaws, but also includes separate oral scan data for the upper jaw and lower jaw.
Since only occlusal data is used in abutment design, to verify whether the upper and lower jaw data alone are useful for pre-training for the abutment design task, we conduct ablation experiments on different methods with different types of pre-training data. 
The experimental results are shown in Table~\ref{Diff_predata}. 
From Table~\ref{Diff_predata}, we can see that adding the separate upper and lower jaw data for network pre-training is beneficial. 
The IoU value of both networks was improved after using the pre-trained model with separate oral scan data.
These experimental results demonstrate that separate upper and lower jaw data can enhance the model’s ability to extract features from oral scan data. 

\begin{table*}
    \centering
    \begin{threeparttable}
        \caption{The IoU performance of different types of pre-training data. UJ: Upper Jaw; LJ: Lower Jaw; OD: Occlusal Data.}
        \begin{tabular}{c|ccc|ccc}
        \hline
        Method & UJ & LJ & OD & Transgingival & Diameter & Height \\ \hline 
        MeshMAE & & & \ding{52} & 40.47 & 64.94  & 30.67 \\
        MeshMAE & \ding{52} & \ding{52} & \ding{52} & \textbf{42.84} & \textbf{65.16} & \textbf{31.52} \\  \hline
        TCEAD & & & \ding{52} & 40.65 & 65.95 & 40.06 \\
        TCEAD & \ding{52} & \ding{52} & \ding{52} & \textbf{43.64} & \textbf{70.78} & \textbf{44.37} \\ \hline
        \end{tabular}
        \begin{tablenotes}
            \small
            \item[] 
        \end{tablenotes}
    \end{threeparttable}
\label{Diff_predata}
\end{table*}

\begin{figure}[htbp]\centering
\centering
\includegraphics[scale=0.25]{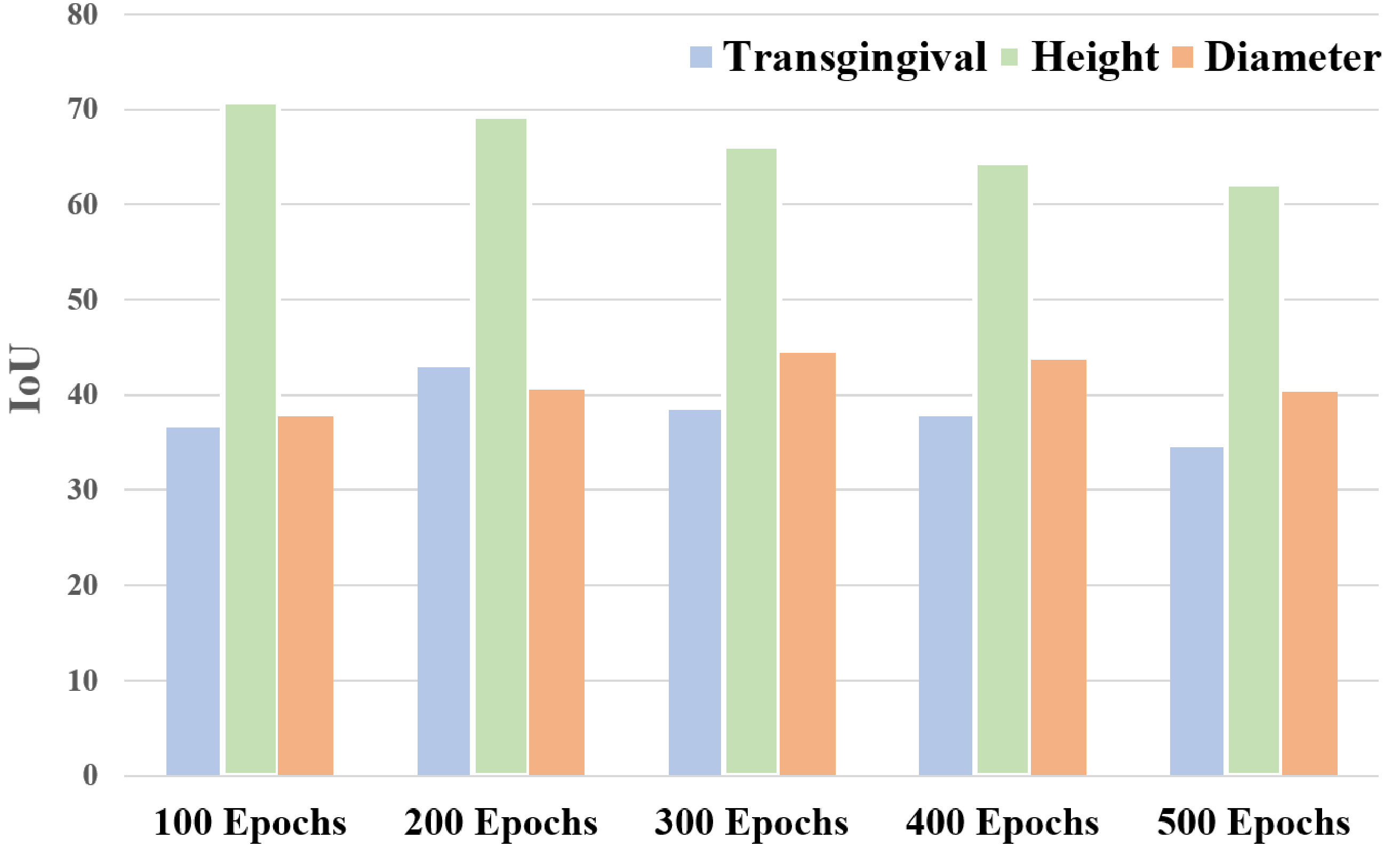}
\caption{The IoU performance under different pre-training epochs.}
\label{diff_epochs}
\end{figure}

\subsubsection{Masking Ratio}
In the pre-training task, the ratio of the mask will affect the network performance. To determine the mask ratio, we conducted comparative experiments with different mask ratios. 
All models were pre-trained for 100 epochs. Experimental results are given in Table~\ref{Different_mask}. 
From Table~\ref{Different_mask}, we can observe that when the mask ratio is 50\%, the model achieves the best IoU in transgingival and diameter. 
When the mask ratio is 60\%, the IoU of height reaches its maximum. 
Experimental results show that different masks will affect the accuracy of the downstream tasks. When the mask ratio is 50\%, our method can effectively complete the regression of the abutment parameter.

\begin{table}[htbp]
\centering
\caption{The IoU performance of different mask ratios in the pre-training stage.}
\begin{tabular}{c|cccc}
\hline
\multirow{2}{*}{Parameter} & \multicolumn{4}{c}{Mask Ratio} \\
  \cline{2-5}
  &  40\% & 50\% & 60\% & 70\% \\ \hline
 Transgingival  & 29.05 &  {42.90} & \textbf{43.95} & 30.40  \\
 Diameter & 66.52 & \textbf{69.31} &  65.93 &  67.19  \\
 Height& 34.05& \textbf{40.59} &  {37.86} &  39.63 \\ \hline
\end{tabular}
\label{Different_mask}
\end{table}

\subsubsection{Different Number of Training Epochs}
We conducted comparative experiments with different epoch numbers to determine the optimal number of pre-training epochs. 
The experimental results are shown in Fig.~\ref{diff_epochs}. 
All experiments were conducted with a 50\% mask ratio. 
As shown in Fig.~\ref{diff_epochs}, the IoU of height gradually decreases with the increase of epoch number, while the IoU of perforation and diameter first increases and then decreases. 
Taking all into consideration, we choose 300 as the number of pre-training epochs.

\begin{figure*}
\centering
\includegraphics[scale=0.23]{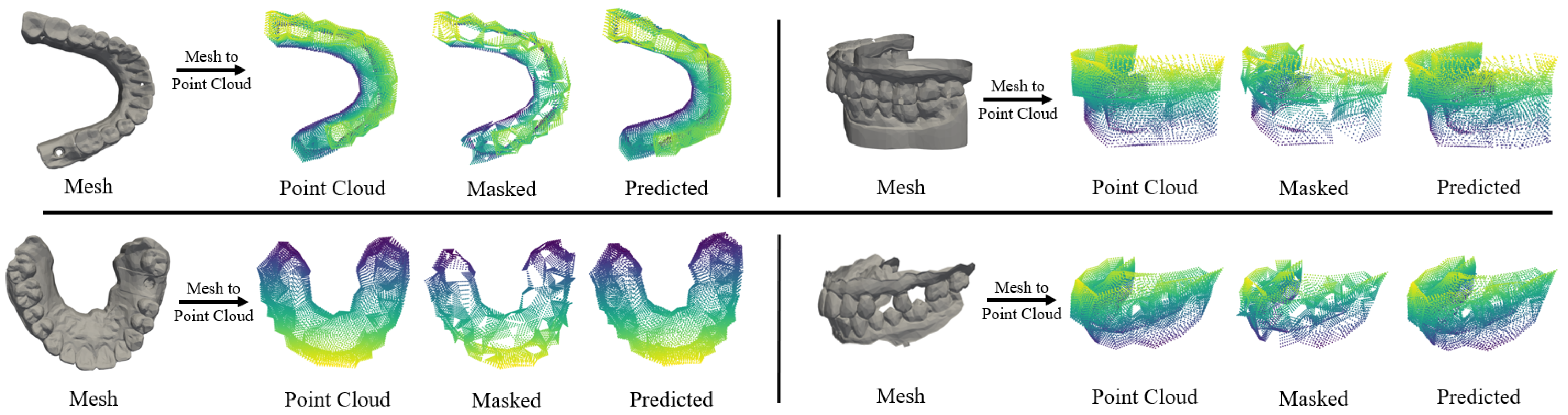}
\caption{Visualization of model reconstruction results. During pre-training, 50\% of the mesh features are randomly masked and subsequently reconstructed.}
\label{Reconstruction}
\end{figure*}

\subsection{Visualization of Reconstruction Results}
To verify whether the pre-trained model has excellent fine-grained feature extraction capabilities, we masked the oral scan data and let the model reconstruct the masked area. The mask ratio was set as 50\%. The visualization result is shown in Fig.~\ref{Reconstruction}.
From the figure, we can see that the texture of the mask area can be well restored, whether it is single jaw or occlusal data, which is close to the original data texture. 
The visualization results demonstrate that our pre-training strategy can greatly improve the feature extraction ability of TCEAD.

\begin{table*}[]
\centering
\begin{threeparttable} \label{Diff_model}
\caption{Compare IoU performance of TCEAD with other mainstream methods.}
\begin{tabular}{c|c|c|ccc}
\hline
	Input & Pre-training & Method & Transgingival & Diameter & Height \\ \hline 
	& \cxmark & PointNet & 26.65 & 36.21 & 21.49 \\
     & \cxmark & PointNet++  & 31.47 & 58.32 & 26.96 \\
	Point Cloud & \cxmark & PointFormer & 29.50& 63.84 & 22.43 \\
     & \ding{52} & PointMAE  & 31.86 & 57.58 & 18.43 \\
     & \ding{52} & PointFEMAE & 33.72 & 59.35 & 19.05 \\ \hline
     & \cxmark & MeshNet++ & 32.01 & 65.05 & 19.51 \\
     Mesh & \ding{52} & MeshMAE & 42.84 & 65.16 & 31.52 \\
	& \ding{52} & TCEAD & \textbf{43.64} & \textbf{70.78} & \textbf{44.37} \\ \hline
	\end{tabular}
        \begin{tablenotes}
            \small
            \item[] 
        \end{tablenotes}
 \end{threeparttable}
\end{table*}

\subsection{Comparison to State-of-the-art Methods}
To further demonstrate the effectiveness of our method, we compare the TCEAD with other mainstream methods, including point cloud-based methods, i.e., PointNet~\citep{qi2017pointnet}, PointNet++~\citep{qi2017pointnet++}, PointFormer~\citep{chen2022pointformer}, PointMAE~\citep{pang2022masked}, PointFEMAE~\citep{zha2024towards}, and mesh-based methods, i.e., MeshNet++~\citep {singh2021meshnet++}, and MeshMAE~\citep{liang2022meshmae}. 
The experimental results are shown in Table~\ref{Diff_model}. 
From the table, we can observe that the TCEAD achieves the best IoU among all parameters. 
Specifically, the mesh-based method outperforms the point cloud-based method in both transgingival and diameter regression tasks. The TCEAD significantly outperforms existing methods in all three regression tasks, particularly in the height regression task, where it achieves a 12\% IoU improvement. 
Compared to methods without pre-training, methods with pre-training are more effective at capturing the features of 3D meshes.

\section{Conclusion}
In this paper, we propose the novel automated implant abutment design solution, a text condition embedded abutment design framework (TCEAD), which fills the gap in the current lack of end-to-end automated design tools for clinical abutment customization. 
Considering that the parameters of the abutment are determined based on the local features of the implant and the opposing tooth, we first pre-trained the model using 7797 sets of oral scan data. This pre-training process enables the model to identify fine-grained features with an IoU improvement of 5.73\%-17.41\% compared to non-pre-trained models.
Moreover, we design a text-guided localization (TGL) module to guide the network to focus on the abutment area, which greatly improves the regression performance.
Extensive experiments are conducted on a large abutment design dataset, and the experimental results demonstrate that TCEAD outperforms existing approaches, achieving an improvement of 0.8\%–12.85\% in IoU compared to mainstream methods.
\par
This paper highlights the potential of TCEAD in reducing the reliance on manual design and improving efficiency in clinical workflows. The proposed framework offers a new direction for intelligent implant-supported restoration, bridging deep learning with practical dental applications. In the future, we plan to validate the method on broader clinical datasets and explore its integration into digital dentistry platforms to further enhance its robustness and clinical usability.

\section*{Acknowledgments}
This work was supported by the National Natural Science Foundation of China under Grant 82261138629; Guangdong Provincial Key Laboratory under Grant 2023B1212060076.

\bibliographystyle{model5-names} 
\bibliography{reference}

\end{document}